\documentclass{article}
\usepackage{spconf,amsmath,graphicx,hyperref}
\usepackage{spconf,amsmath,graphicx,hyperref}

\usepackage[utf8]{inputenc} 
\usepackage[T1]{fontenc}    
\usepackage{url}            
\usepackage{booktabs}       
\usepackage{amsfonts}       
\usepackage{nicefrac}       
\usepackage{microtype}      
\usepackage{xcolor}         

\usepackage[accsupp]{axessibility}  

\usepackage{cite}
\usepackage{caption}
\usepackage{subcaption}
\usepackage{graphicx}
\usepackage{hyperref}
\usepackage{cleveref}

\usepackage{lipsum}
\usepackage{wrapfig}
\usepackage{array}
\usepackage{xcolor}
\usepackage{makecell}
\usepackage{enumitem}

\usepackage{algorithm}
\usepackage{algpseudocode}
\floatname{algorithm}{Algorithm}
\algrenewcommand\algorithmicrequire{\textbf{Require:}}
\algrenewcommand\algorithmicensure{\textbf{Ensure:}}

\usepackage{orcidlink}
\usepackage[dvipsnames]{xcolor}
\usepackage{booktabs}

\usepackage{pifont} 

\usepackage{colortbl}
\usepackage{multirow}
\usepackage{multicol}
\usepackage{listings} 
\usepackage{titletoc} 
\usepackage{fancyvrb}
\usepackage{tcolorbox} 
\usepackage[accsupp]{axessibility}  
\usepackage[subtle]{savetrees} 

\definecolor{lightgray}{rgb}{0.83, 0.83, 0.83}
\definecolor{Gray}{gray}{0.6}
\definecolor{aliceblue}{rgb}{0.94, 0.97, 1.0}
\definecolor{mistyrose}{rgb}{1.0, 0.89, 0.88}
\definecolor{backcolour}{rgb}{0.95,0.95,0.92}

\newcommand{\appendixref}[2]{%
  \if\sepappendix1%
    #1
  \else%
    #2
  \fi%
}

\usepackage{cleveref}

\crefname{equation}{Eq.}{Eqs.}
\Crefname{equation}{Equation}{Equations}

\crefname{figure}{Fig.}{Figs.}
\Crefname{figure}{Figure}{Figures}

\crefname{table}{Tab.}{Tabs.}
\Crefname{table}{Table}{Tables}

\crefname{section}{Sec.}{Secs.}
\Crefname{section}{Section}{Sections}

\crefname{algorithm}{Alg.}{Algs.}
\Crefname{algorithm}{Algorithm}{Algorithms}

\usepackage{etoolbox}
\AtBeginDocument{%
  \setlength{\abovedisplayskip}{3pt}%
  \setlength{\belowdisplayskip}{3pt}%
  \setlength{\abovedisplayshortskip}{2pt}%
  \setlength{\belowdisplayshortskip}{2pt}%
}

\title{Tracing Audio Grounding and Answer Selection in Audio LLMs}

\name{
  Hyebin Cho \quad
  Suho Yoo \quad
  Jihoo Jung \quad
  Joon Son Chung
}
\address{
Korea Advanced Institute of Science and Technology, South Korea \\
}

\begin{document}
%
\maketitle

\begin{abstract}
Audio Large Language Models (Audio LLMs) have advanced in audio understanding, yet they can still predict the answer by reasoning from textual cues or linguistic priors rather than the provided audio. 
A common remedy is to train models on data whose answers cannot be inferred from text alone. This approach can improve performance, but what changes within the model remains unclear.
In this paper, we ask what must happen inside the model for the audio to actually determine the answer. 
Our findings are threefold.
(1) Replacing the audio with silence or unrelated audio causes substantially larger performance degradation in the trained model than in the pretrained model.
(2) Acoustic information most strongly shapes the model's representations of the answer choices in early-to-middle layers, while training mainly increases the influence of audio information on the final prediction in middle-to-late layers.
(3) The weights learned during training have their largest impact in specific layer bands.
Together, these results provide a mechanistic account of how training strengthens the use of acoustic evidence in Audio LLMs.
\end{abstract}
\vspace{-1.5em}

\section{Introduction}

Recent Audio Large Language Models (Audio LLMs) have substantially advanced audio understanding and reasoning. However, beyond task accuracy, recent evaluation increasingly asks whether model predictions are actually supported by the acoustic evidence in the input, particularly under hallucination, modality conflict, missing evidence, and other challenging conditions~\cite{kuan2025can,kuan2024understanding,yang2026multi,kuan2026aqua,li2026silence,yin2026can,seth2026exploring}. Furthermore, recent work shows that high AudioQA performance can persist even when little or no acoustic evidence is available, highlighting a gap between benchmark accuracy and the actual use of acoustic evidence~\cite{foo2026all}.

In response to this limitation, one line of work has developed training data and evaluation benchmarks in which correct answers require information from the audio~\cite{AudioMCQ,he2026summary}. In parallel, mechanistic studies have begun to examine how acoustic information is represented and propagated within Audio LLMs, including where audio-related information is represented and how audio and text interact~\cite{yang2025audiolens,zhao2026discovering,glazer2026listening,cho2026wins,ma2026semantics,chen2026causal}. 
However, it remains unclear how training changes the way audio information is used inside an Audio LLM to produce the final answer.

\begin{figure}[t]
    \centering
    \includegraphics[width=\columnwidth]{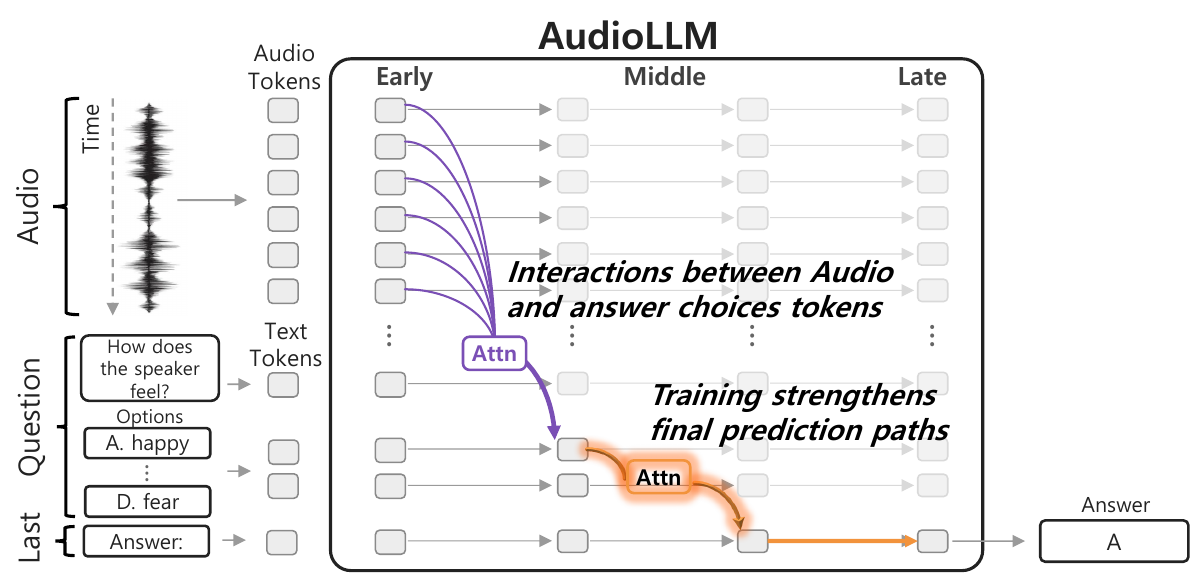}
    \vspace{-1.5em}
    \caption{\textbf{How audio information is used across layers in Audio LLMs.}
    Connections from audio to the answer choices are most important in early-to-middle layers both before and after training, while training increases the contribution of the answer choices to the final prediction in middle-to-late layers.}
     \label{fig:teaser}
    \vspace{-1.75em}
\end{figure}

We therefore ask three questions. 
First, does training on questions that require acoustic evidence increase the model's use of the provided audio? 
Second, at which layers does acoustic information shape the model's representations of the answer choices, and at which layers do these representations begin to influence the final prediction?
Third, where are the learned changes that support these improvements concentrated?

We study these questions across multiple-choice AudioQA benchmarks, analyzing both changes in model behavior and where those changes arise within the model. \cref{fig:teaser} summarizes the resulting picture. Our main findings are:


\vspace{-0.3em}
\begin{itemize}
\setlength{\itemsep}{1pt}
\setlength{\parskip}{0pt}
\setlength{\parsep}{0pt}

\item \textbf{Training increases the use of acoustic evidence:} on benchmarks where training improves performance, replacing the audio with silence or unrelated audio causes larger accuracy drops after training than before training.

\item \textbf{Acoustic information shapes the answer choices before influencing the prediction:}
it most strongly shapes answer-choice representations in early-to-middle layers, whereas training strengthens their influence on the final prediction in middle-to-late layers.

\item \textbf{The effects of learned parameter updates are layer-specific:}
disabling these updates causes the largest performance drops in model-specific middle-to-late or late layer ranges.
\end{itemize}
\vspace{-0.3em}

Overall, our results provide a mechanistic view of where and how training strengthens the use of acoustic evidence in Audio LLMs, distinguishing acoustic information integration from its contribution to final answer selection.


\section{Method}

\subsection{Analysis Setup}

\noindent\textbf{Task and data.}
We study multiple-choice AudioQA, where each example consists of an audio input, a question, a set of candidate answers, and a correct option label. Given the audio, question, and answer choices, the model is instructed to generate only the corresponding option letter.

For training, we use the AudioMCQ-StrongAC-GeminiCoT dataset ($19{,}480$ samples)~\cite{AudioMCQ}, the DCASE 2026 training set curated to favor questions that require acoustic evidence for correct answering. We discard the chain-of-thought annotations and use only the option letter for supervision.

We evaluate cross-benchmark generalization on ADQA-Bench ($2{,}998$ samples)~\cite{ADQA}, MMAU-test-mini ($1{,}000$ samples)~\cite{mmau}, MMAR ($1{,}000$ samples)~\cite{mmar}, and MMSU ($5{,}000$ samples)~\cite{mmsu}, none of which is used during training. ADQA-Bench is the official DCASE 2026 evaluation set and is specifically constructed to reduce questions that can be answered from textual cues alone, providing a targeted evaluation of acoustic-evidence use.

\noindent\textbf{Models and fine-tuning.}
We analyze Qwen2-Audio~\cite{qwen2} and Qwen2.5-Omni~\cite{qwen25omni}, comparing their original instruction-tuned checkpoints, denoted as zero-shot (\textsc{ZS}), with their LoRA-adapted counterparts, denoted as \textsc{FT}. Training uses four answer choices per question, whereas evaluation retains the original number of choices for each sample and restricts generation to the corresponding valid option labels. We apply LoRA with rank $r=16$, scaling factor $\alpha=32$, and dropout $0.05$ to the audio encoder and language-model target modules, with input audio truncated to at most 30 seconds.

\subsection{Behavioral Sensitivity to Acoustic Evidence}

To assess how training changes the model's sensitivity to relevant acoustic evidence, we evaluate each model under three audio conditions. In the \textit{original} condition, the paired audio is provided without modification. In the \textit{silent} condition, the waveform is replaced with an all-zero signal of the same duration, removing informative acoustic content while preserving the audio input. In the \textit{mismatch} condition, the question and answer choices are retained, while the audio is replaced with audio from another randomly selected sample in the same benchmark. This preserves a natural waveform while breaking its correspondence with the question. We construct mismatched pairs using a fixed random seed and exclude identity matches.

Let $A_{\mathrm{orig}}$, $A_{\mathrm{sil}}$, and $A_{\mathrm{mis}}$ denote the accuracies under the original, silent, and mismatched conditions, respectively. We average the silent and mismatched conditions to reduce dependence on either perturbation alone. We define the behavioral acoustic-evidence sensitivity as
\begin{equation}
S_{\mathrm{audio}}=A_{\mathrm{orig}}-\frac{A_{\mathrm{sil}} + A_{\mathrm{mis}}}{2}.
\end{equation}
A larger $S_{\mathrm{audio}}$ indicates greater performance degradation when valid acoustic evidence is unavailable. We measure the change after training as
\begin{equation}
\Delta S_{\mathrm{audio}}=S_{\mathrm{audio}}^{\mathrm{FT}}-S_{\mathrm{audio}}^{\mathrm{ZS}},
\end{equation}
where a positive value indicates increased behavioral sensitivity to valid acoustic evidence. This score measures behavioral sensitivity rather than directly identifying the internal mechanism by which audio influences the prediction.


\vspace{-1em}
\subsection{Layer-Wise Attention Interventions}

We use attention knockout~\cite{geva2023dissecting,kim2025map} to
identify attention connections that are functionally important for
prediction. During inference, we prevent a predefined target token
group from attending to a source token group within selected layers,
while keeping all model parameters unchanged.

We partition the input into audio tokens $\mathcal{T}_{a}$, question
tokens $\mathcal{T}_{q}$, candidate-answer tokens
$\mathcal{T}_{o}$, and the first generated answer position
$\mathcal{T}_{y}$. For source tokens $S$ and target tokens $T$, we
modify the pre-softmax attention score as
\begin{equation}
\widetilde z_{t,s}^{\ell,h}
=
\begin{cases}
-\infty,
& t\in T,\;s\in S,\;\ell\in\mathcal{W},\\
z_{t,s}^{\ell,h},
& \text{otherwise},
\end{cases}
\end{equation}
where $\mathcal{W}$ denotes the intervened layer window. This sets the corresponding attention weights to zero after softmax, preventing tokens in $T$ from attending to tokens in $S$ across all attention heads. We denote this intervention as $S\nrightarrow T$.

We focus on two connections representing different stages of multiple-choice prediction.  \textit{Audio$\nrightarrow$Options} prevents candidate-answer tokens from attending to the audio and tests the functional importance of audio-to-option interactions.
\textit{Options$\nrightarrow$Answer} prevents the generated answer position from attending to the candidate answers and tests their direct contribution to the final prediction.

For each intervention, we measure the mean change in the probability assigned to the correct option after renormalizing the next-token probabilities over the valid answer labels for each sample:
\begin{equation}
\Delta p^{S\nrightarrow T}_{\ell}=\frac{100}{N}\sum_{i=1}^{N}\left[p^{\mathrm{KO}}_{i,\ell}(y_i)-p^{\mathrm{full}}_{i}(y_i)\right].
\end{equation}
A negative $\Delta p$ indicates that removing the connection reduces support for the correct answer, with larger negative values indicating a stronger functional contribution. We use probability rather than argmax accuracy to capture intervention effects that may change model confidence without changing the predicted option. For the layer-wise analyses, we pool all examples from the benchmarks and compute $\Delta p$ over the pooled examples.

We center each $k$-layer intervention window at layer $\ell$, spanning $\ell-(k-1)/2$ through $\ell+(k-1)/2$ and clipping the window at model boundaries. We use $k=9$ for the main analysis and additionally evaluate $k\in\{1,3,5,9,13\}$, observing qualitatively consistent layer-wise trends.

\subsection{Layer-wise Analysis of Learned Updates}

To identify where training-induced changes have the largest functional effect, we group the learned LoRA updates into contiguous layer bands across the jointly adapted audio and language modules. We first disable one trained band at a time while keeping the remaining adapter unchanged. We then train separate adapters restricted to a single band and compare them with full-depth LoRA. We additionally include full-depth LoRA with $r=4$, which approximately matches the trainable-parameter budget of single-band $r=16$ training.
\begin{table}[t]
\centering
\caption{\textbf{Accuracy and acoustic sensitivity.} Accuracies are in percentages. $\Delta A$ and $\Delta S_{\mathrm{audio}}$ denote changes in accuracy and behavioral sensitivity to valid acoustic evidence, respectively, in percentage points. $\pm$ values indicate 95\% paired-bootstrap CI half-widths from 10,000 resamples.}
\label{tab:audio_dependence_v2}
\vspace{-4pt}
\scriptsize
\setlength{\tabcolsep}{15pt}\small
\resizebox{\columnwidth}{!}{%
\begin{tabular}{@{}lcccc@{}}
\toprule
\textbf{Dataset}
& \textbf{ZS}
& \textbf{FT}
& \(\boldsymbol{\Delta A}\)
& \(\boldsymbol{\Delta S_{\mathrm{audio}}}\) \\
\midrule

\multicolumn{5}{@{}c}{\textit{Qwen2-Audio}} \\[-1pt]
\midrule
ADQA
& 36.9
& 51.1
& $+14.3{\pm}2.0$
& $+9.3{\pm}2.0$ \\

MMAR
& 41.2
& 52.5
& $+11.3{\pm}3.9$
& $+6.9{\pm}3.6$ \\

MMAU
& 59.7
& 58.6
& $-1.1{\pm}3.6$
& $-9.1{\pm}3.5$ \\

MMSU
& 47.3
& 58.8
& $+11.5{\pm}1.6$
& $+6.8{\pm}1.5$ \\

\midrule
\multicolumn{5}{@{}c}{\textit{Qwen2.5-Omni}} \\[-1pt]
\midrule
ADQA
& 41.9
& 54.4
& $+12.5{\pm}1.9$
& $+4.5{\pm}2.0$ \\

MMAR
& 53.7
& 65.7
& $+11.9{\pm}3.4$
& $+6.7{\pm}3.5$ \\

MMAU
& 77.0
& 63.5
& $-13.5{\pm}3.2$
& $-7.1{\pm}3.1$ \\

MMSU
& 54.1
& 64.9
& $+10.9{\pm}1.4$
& $+6.0{\pm}1.5$ \\

\bottomrule
\end{tabular}
}
\vspace{-10pt}
\end{table}

\section{Experiments}

\subsection{Behavioral Use of Acoustic Evidence}

Training improves accuracy on ADQA-Bench, MMAR, and MMSU for both models, while MMAU shows little improvement for Qwen2-Audio and a clear decrease for Qwen2.5-Omni (\cref{tab:audio_dependence_v2}).

We next ask whether these performance gains are accompanied by greater use of the provided audio. We operationalize behavioral audio use as sensitivity to removing or replacing the paired audio. On all six model--dataset pairs where accuracy improves, $\Delta S_{\mathrm{audio}}$ is also positive, showing that the trained models become more sensitive to removing or replacing the corresponding audio. In contrast, both models show negative $\Delta S_{\mathrm{audio}}$ on MMAU, together with weak or negative accuracy changes. These results suggest that successful performance transfer is generally accompanied by a greater contribution from relevant acoustic evidence, whereas the learned changes do not generalize in the same way to MMAU.

Because the pathway analysis targets settings where training improves both performance and audio sensitivity, we focus on ADQA-Bench, MMAR, and MMSU; MMAU is retained as a negative-transfer comparison in the learned-update analysis.

\begin{figure}[t]
    \centering
    \includegraphics[width=\columnwidth]{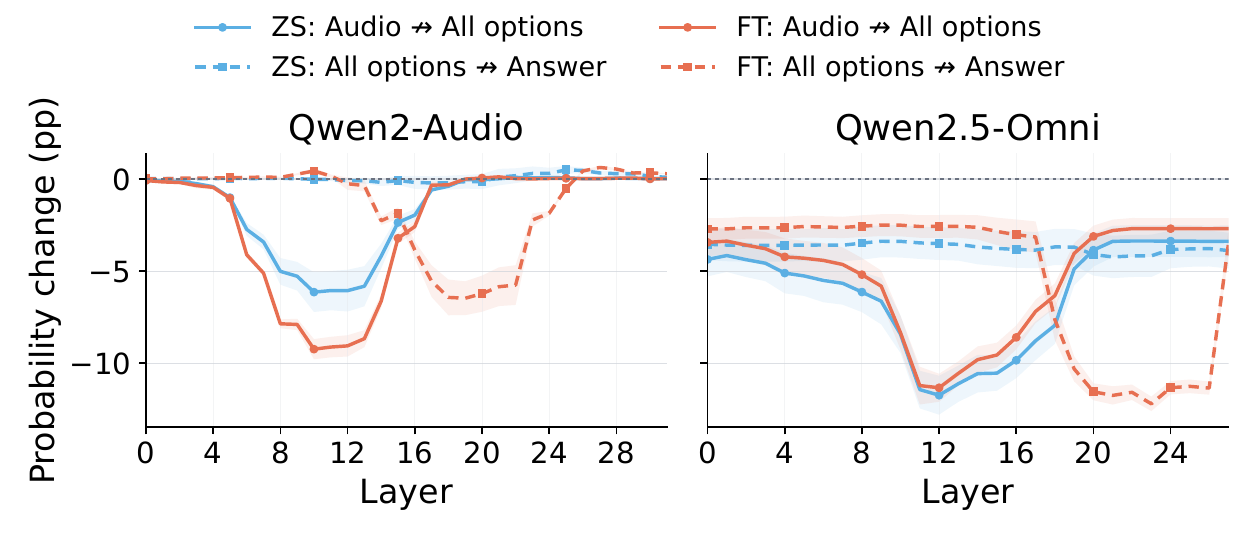}
    \vspace{-2.5em}
    \caption{\textbf{Layer-wise attention knockout.}
    Audio$\nrightarrow$All Options is most influential in early-to-middle layers for both \textsc{ZS} and \textsc{FT}, whereas All Options$\nrightarrow$Answer becomes prominent in middle-to-late layers after training.}
    \label{fig:average_knockout}
    \vspace{-0.75em}
\end{figure}

\begin{figure}[t]
    \centering
    \includegraphics[width=\columnwidth]{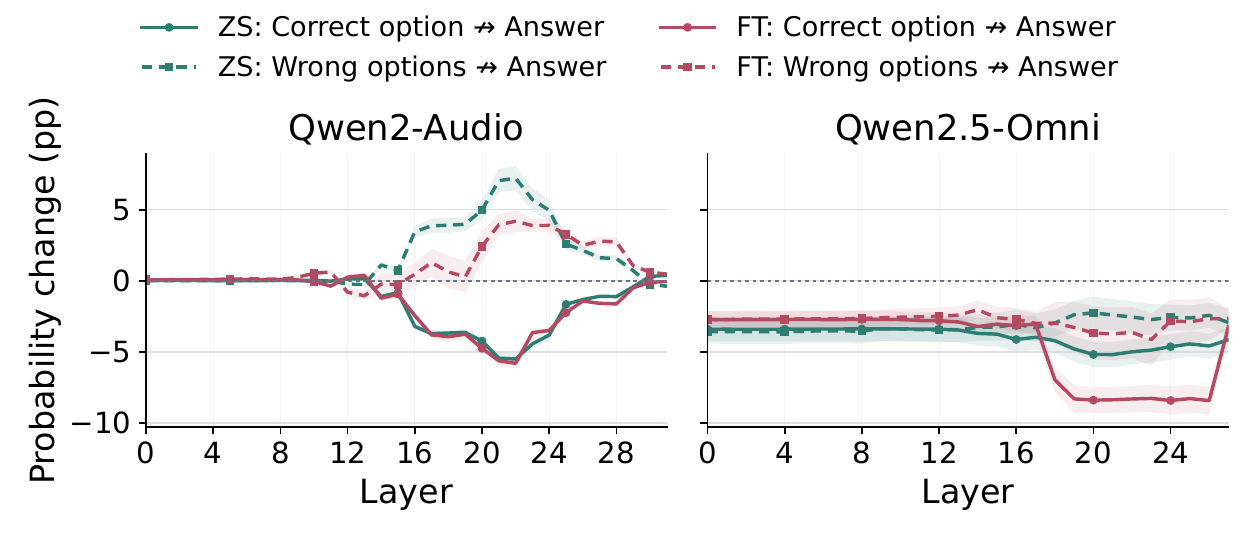}
    \vspace{-2.5em}
    \caption{\textbf{Correct- and wrong-option contributions.} Qwen2-Audio shows opposite effects for correct and wrong options in both \textsc{ZS} and \textsc{FT}. Blocking the correct option lowers the correct-answer probability, while blocking wrong options raises it. Qwen2.5-Omni shows a strong increase in correct-option contribution after training, with little change from wrong options.}
    \label{fig:option_knockout}
    \vspace{-0.75em}
\end{figure}

\subsection{From Acoustic Evidence to Answer Prediction}

\cref{fig:average_knockout} shows a similar layer-wise pattern in both Qwen2-Audio and Qwen2.5-Omni. Blocking Audio$\nrightarrow$All Options reduces the probability of the correct answer in early-to-middle layers for both \textsc{ZS} and \textsc{FT}. The strongest effects appear around layers 8--12 in Qwen2-Audio and 10--16 in Qwen2.5-Omni. This indicates that acoustic information influences the representations of the candidate answers primarily in early-to-middle layers, and that this effect is already present before training.

A different pattern appears for All Options$\nrightarrow$Answer. In the zero-shot models, blocking this connection produces little change across layers. After training, however, the same intervention causes a clear reduction in the correct-answer probability at later layers, around 18--22 in Qwen2-Audio and 20--27 in Qwen2.5-Omni. Thus, the largest change after training is not in the earlier interaction between audio and candidate answers, but in how those representations contribute to the final prediction.

Overall, acoustic evidence affects candidate-answer representations earlier, whereas their contribution to the final prediction becomes prominent later. Training mainly strengthens this later use of audio-informed representations. As a result, removing or replacing the audio has a larger effect on the final prediction, even though the earlier acoustic-grounding stage itself changes less.

To determine whether the strengthened option-to-answer connection reflects general dependence on the answer choices or more selective use of the correct option, we further separate the contributions of the correct and incorrect options in \cref{fig:option_knockout}. In Qwen2-Audio, blocking Correct Option$\nrightarrow$Answer decreases the correct-answer probability over approximately layers 16--28, whereas blocking Wrong Options$\nrightarrow$Answer increases it. This pattern appears in both \textsc{ZS} and \textsc{FT}, indicating that the model already distinguishes between supportive information from the correct option and competing information from the incorrect options before training.

Qwen2.5-Omni shows a different training-induced change. Blocking the wrong options has little effect in either setting, whereas blocking Correct Option$\nrightarrow$Answer produces a pronounced decrease in the correct-answer probability after training, particularly around layers 18--25. This suggests that training primarily strengthens the contribution of the correct-option representation to the final answer in Qwen2.5-Omni.

\begin{figure}[t]
    \centering
    \includegraphics[width=\columnwidth]{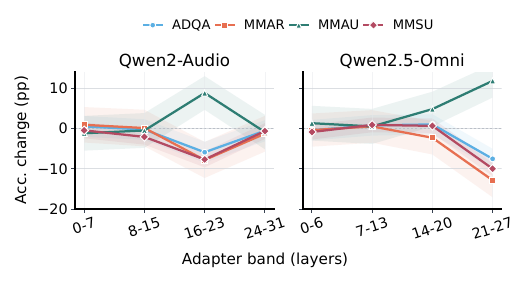}
    \vspace{-1.5em}
    \caption{\textbf{Functional effect of LoRA layer bands.} Accuracy change after disabling the learned LoRA updates in each layer band, relative to the full trained model. Negative values indicate performance degradation after removal, whereas positive values indicate improved accuracy.}
    \label{fig:lora_band_ablation}
    \vspace{-0.75em}
\end{figure}

\subsection{Where Learned LoRA Updates Matter}

We next ask where the learned LoRA updates have the largest functional effect, and whether the same layer ranges are sufficient when used alone.

\noindent\textbf{Functional necessity.} We first disable the learned LoRA updates in one layer band at a time while keeping the rest of the trained model unchanged. As shown in \cref{fig:lora_band_ablation}, removing layers 16--23 causes the largest performance drop in Qwen2-Audio, while Qwen2.5-Omni is most affected by removing layers 21--27. These ranges overlap with the layers where Options$\nrightarrow$Answer becomes more important after training in \cref{fig:average_knockout}. This suggests that the learned changes are concentrated in particular layer ranges rather than being equally important across all layers. MMAU provides a complementary negative-transfer case: disabling some learned bands improves accuracy, consistent with the weak or negative transfer observed in \cref{tab:audio_dependence_v2}.

\noindent\textbf{Adaptation sufficiency.} We next train LoRA while allowing updates in only one layer band at a time. In \cref{tab:layerband_lora_ft}, these single-band models use rank $r=16$ and are compared with all-layer LoRA using $r=16$ and $r=4$. For Qwen2-Audio, training only layers 8--15 nearly matches or exceeds the all-layer $r=16$ model on ADQA-Bench, MMAR, and MMSU, despite updating only one quarter of the layer locations. Qwen2.5-Omni behaves differently: no single band recovers the performance of all-layer training, and even the all-layer $r=4$ model performs better on average than any single-band model. This suggests that Qwen2-Audio can acquire much of the improvement from a restricted layer range, whereas Qwen2.5-Omni benefits from learned updates across multiple layer ranges.

The band that matters most to the fully trained model is not necessarily the band that performs best when trained in isolation. Band ablation asks which learned updates the full model relies on after training, whereas single-band training asks what can be learned when updates are restricted from the beginning. This difference suggests that learned updates across layer bands can work together during training and inference.

\begin{table}[t]
\centering
\caption{\textbf{Layer-restricted LoRA training.} Accuracy (\%) when LoRA updates are applied to all layers or restricted to a single layer band during training. Unless otherwise specified, LoRA rank is $r=16$. Bold and underline indicate the best and second-best results in each row, respectively.}
\label{tab:layerband_lora_ft}
\vspace{-4pt}
\scriptsize
\setlength{\tabcolsep}{8pt}\small
\resizebox{\columnwidth}{!}{%
\begin{tabular}{@{}lrrrrrr@{}}
\toprule
\multicolumn{7}{@{}c}{\textit{Qwen2-Audio}} \\
\midrule
\textbf{Data} & \textbf{All} & \textbf{0--7} & \textbf{8--15} & \textbf{16--23} & \textbf{24--31} & \textbf{All $r=4$} \\
\midrule
ADQA
& $\mathbf{51.13}$
& $\underline{49.73}$
& $\mathbf{51.13}$
& $38.80$
& $41.60$ 
& $39.70$ \\

MMAR
& $\underline{52.51}$
& $49.50$
& $\mathbf{52.81}$
& $42.11$
& $44.92$ 
& $43.82$ \\

MMAU
& $\mathbf{58.60}$
& $56.60$
& $55.60$
& $55.90$
& $57.50$ 
& $\underline{57.90}$ \\

MMSU
& $\underline{58.78}$
& $56.68$
& $\mathbf{59.72}$
& $51.34$
& $52.56$ 
& $51.58$ \\

\midrule
Mean
& $\mathbf{55.26}$
& $53.13$
& $\underline{54.82}$
& $47.04$
& $49.15$ 
& $48.25$
\\

\midrule
\multicolumn{7}{@{}c}{\textit{Qwen2.5-Omni}} \\
\midrule
\textbf{Data} & \textbf{All} & \textbf{0--6} & \textbf{7--13} & \textbf{14--20} & \textbf{21--27} & \textbf{All $r=4$} \\
\midrule
ADQA
& $\mathbf{54.37}$
& $41.10$
& $41.17$
& $39.97$
& $40.47$ 
& $\underline{42.00}$
\\

MMAR
& $\mathbf{65.66}$
& $53.27$
& $56.38$
& $54.87$
& $51.36$ 
& $\underline{56.88}$
\\

MMAU
& $63.50$
& $63.00$
& $\mathbf{64.60}$
& $61.90$
& $62.50$ 
& $\underline{64.40}$
\\

MMSU
& $\mathbf{64.94}$
& $54.30$
& $\underline{55.98}$
& $53.44$
& $53.34$ 
& $55.38$
\\

\midrule
Mean
& $\mathbf{62.12}$
& $52.92$
& $54.53$
& $52.55$
& $51.92$ 
& $\underline{54.67}$ 
\\

\bottomrule
\end{tabular}
}
\vspace{-8pt}
\end{table}





\section{Conclusion}
We studied how training changes the use of acoustic evidence in Audio LLMs. Acoustic evidence affects candidate-answer representations in early-to-middle layers, while training mainly strengthens their contribution to the final answer in middle-to-late layers, with learned effects concentrated in model-specific LoRA bands. Our results highlight an important distinction in Audio LLMs: making acoustic evidence available to answer representations and using those representations for the final decision are distinct stages.

\begingroup

\apptocmd{\thebibliography}{
  \setlength{\itemsep}{0pt}
  \setlength{\parskip}{0pt}
}{}{}

\bibliographystyle{IEEEbib}
\bibliography{shortstrings, refs}

\end{document}